\documentclass[12pt,peerreviewca,onecolumn,draftcls]{IEEEtran}
\newcommand{\FigDir}{tinyfigs}
\usepackage{edsstuff}
\usepackage[pagebackref]{hyperref}
\usepackage[sort]{cite}
\usepackage{url}

\def\infig#1{\textsf{#1}}

\begin{document}

\def\Eqn#1{(\ref{eqn:#1})}

\title{Faster and better: a machine learning approach to corner detection}
\author{Edward~Rosten, Reid~Porter, and Tom~Drummond%
\thanks{Edward Rosten and Reid Porter are with Los Alamos National Laboratory, Los Alamos, New Mexico, USA, 87544.
Email: edrosten@lanl.gov, rporter@lanl.gov}%
\thanks{Tom Drummond is with Cambridge University, Cambridge University Engineering Department, Trumpington Street, Cambridge, UK,  CB2 1PZ
Email: twd20@cam.ac.uk}}
\maketitle

\begin{abstract}

The repeatability and efficiency of a corner detector determines how likely it is to be useful in a real-world application. The repeatability is importand because the same scene viewed from different positions should yield features which correspond to the same real-world 3D locations~\cite{schmid2000comparing}.  The efficiency is important because this determines whether the detector combined with further processing can operate at frame rate.

Three advances are described in this paper.  First, we present a new heuristic for feature detection, and using machine learning we derive a feature detector from this which can fully process live PAL video using {\em less than 5\%} of the available processing time.  By comparison, most other detectors cannot even operate at frame rate (Harris detector 115\%, SIFT 195\%).  Second, we generalize the detector, allowing it to be optimized for repeatability, with little loss of efficiency.  Third, we carry out a rigorous comparison of corner detectors based on the above repeatability criterion applied to 3D scenes.  We show that despite being principally constructed for speed, on these stringent tests, our heuristic detector significantly outperforms existing feature detectors.  Finally, the comparison demonstrates that using machine learning produces significant improvements in repeatability, yielding a detector that is both very fast and very high quality.

\end{abstract} 

\def\CornerResponse{\ensuremath{C}\xspace}
\def\Image{\ensuremath{I}\xspace}

\section{Introduction}

\PARstart{C}{orner} detection is used as the first step of many vision tasks such as
tracking, localisation, SLAM (simultaneous localisation and mapping), image
matching and recognition.  This need has driven the development of a large
number of corner detectors.  However, despite the massive  increase in
computing power since the inception of corner detectors, it is still true that
when processing live video streams at full frame rate, existing feature
detectors leave little if any time for further processing.

In the applications described above, corners are typically detected and matched
into a database, thus it is important that the same real-world points are
detected repeatably from multiple views \cite{schmid2000comparing}. The amount
of variation in viewpoint under which this condition should hold depends on the
application.  

\section{Previous work}
\label{sec:lit}
\subsection{Corner detectors}

Here we review the literature to place our advances in
context.  
In the literature, the terms ``point feature'', ``feature'', ``interest point'' and
``corner'' refer to a small 
point of interest with variation in two dimensions.
Such points often arise as the result of geometric
discontinuities, such as the corners of real world objects, but they may also
arise from small patches of texture. Most algorithms are capable of detecting
both kinds of points of interest, though the algorithms are often  designed to
detect one type or the other.
A number of the detectors described below compute a corner response,
\CornerResponse, and define corners to be large local maxima of \CornerResponse.

%
%

\subsubsection{Edge based corner detectors}

An edge (usually a step change in intensity) in an image corresponds to the
boundary between two regions.  At corners, this boundary changes direction
rapidly.

\paragraph{Chained edge based corner detectors}

Many techniques have been developed which involved detecting 
and chaining edges with a view to analysing the properties of the
edge, often taking points of high curvature to be corners.  Many early methods
used chained curves, and since the curves are highly quantized, the techniques
concentrate on methods for effectively and efficiently estimating the
curvature.  A common approach has been to use a chord for estimating the slope of a
curve or a pair of chords to find the angle of the curve at a point.

Early methods computed the smallest angle of the curve over chords spanning
different numbers of links.  Corners are defined as local minima of
angle~\cite{rosenfeld73angle} after local
averaging~\cite{rosenfeld75improved}. Alternatively, corners can be defined as
isolated discontinuities in the mean slope, which can be computed using a chord spanning a
fixed set of links in the chain~\cite{freeman77corner}. Averaging can be used to
compute the slope and the length of the curve used to determine if a
point is isolated~\cite{beus87improved}. The angle can be computed using a pair
of chords with a central gap, and peaks with certain widths (found by looking
for zero crossings of the angle) are defined as corners~\cite{ogorman88feature}.

Instead of using a fixed set of chord spans, some methods compute a `region of
support' which depends on local curve properties. For instance local maxima of
chord lengths can be used to define the region of support, within which a corner
must have maximal curvature~\cite{teh89dominant}.  Corners are can be defined as
the centre of a region of support with high mean curvature, where the support
region is large and symmetric about its centre~\cite{ogawa89corner}.  The region
free from significant discontinuities around the candidate point can be used
with curvature being computed as the slope change across the
region~\cite{bandera00corner} or the angle to the region's
endpoints~\cite{urdiales03corner}.

An alternative to using chords of the curves is to apply smoothing to the points
on the curve.  Corners can be defined as points with a high rate of change of
slope~\cite{sohn92corner},  or points
where the curvature decreases rapidly to the nearest minima and the angle to the
neighbouring maxima is small~\cite{he04corner}. 

A fixed smoothing scale is not necessarily appropriate for all curves, so
corners can also be detected at high curvature points which have stable positions
under a range of smoothing scales~\cite{ansari91detecting}.  As smoothing is
decreased, curvature maxima bifurcate, forming a tree over scale. Branches of a
tree which are longer (in scale) than the parent branch are considered as stable
corner points~\cite{chin92corner}. Instead of Gaussian smoothing, extrema of the
wavelet transforms of the slope~\cite{lee92corner} or wavelet transform modulus
maximum of the angle~\cite{lee95corner,quddus99corner} over multiple scales can
be taken to be corners.

The smoothing scale can be chosen adaptively.  The Curvature Scale Space
technique~\cite{CSS} uses a scale proportional to the length and defines corners at
maxima of curvature where the maxima are significantly larger than the closest
minima. Locally adaptive smoothing using anisotropic
diffusion~\cite{medioni91adaptive} or smoothing scaled by the local variance of
curvature~\cite{ray03corner} have also been proposed.

Instead of direct smoothing, edges can be parameterised with cubic splines and
corners detected at points of high second derivative where the spline deviates
a long way from the control point~\cite{langridge-corner,medioni-corner}.

A different approach is to extend curves past the endpoints by following saddle
minima or ridge maxima in the gradient image until a nearby edge is crossed,
thereby finding junctions~\cite{beymer91junctions}.  Since the chain code number
corresponds roughly to slope, approximate curvature can be found using finite
differences, and corners can be found by identifying specific
patterns~\cite{seeger94corner}.  Histograms of the chain code numbers on either
side of the candidate point can be compared using normalized cross correlation
and corners can be found at small local minima~\cite{arrebola97corner}. Also, a measure
of the slope can be computed using circularly smoothed histograms of the chain
code numbers~\cite{arrebola99corner}.  Points can be classified as corners using
fuzzy rules applied to measures computed from the forward and backward arm and
the curve angle~\cite{li99corner}.

%

\paragraph{Edgel based corner detectors}

Chained edge techniques rely on the method used to perform segmentation and edge
chaining, so many techniques find edge points (edgels) and examine the local
edgels or image to find corners.

For instance, each combination of presence or absence  of edgels in a $3\times3$
window can be assigned a curvature, and corners found as maxima of curvature in
a local window~\cite{sankar78parallel}. Corners can be also found by analysing
edge properties in the window scanned along the edge~\cite{cheng88corner}.
A generalized Hough transform~\cite{ballard81hough} can be used which replaces
each edgel with a line segment, and corners
can be found where lines intersect,
\ie at large maxima in Hough space~\cite{davies88corner}.  In a manner similar to
chaining, a short line segment can be fitted to the edgels, and the corner
strength found by the change in gradient direction along the line
segment~\cite{computer-and-robot-vision}. Edge detectors often fail at
junctions, so corners can be defined as points where several edges at different
angles end nearby~\cite{mehrotra90corner}. By finding both edges and their
directions, a patch on an edge can be compared to patches on either side in the
direction of the contour, to find points with low
self-similarity~\cite{cooper91corner}.

Rapid changes in the edge direction can be found by measuring the derivative of the
gradient direction along an edge and multiplying  by the magnitude of the
gradient:
\begin{align}
\label{eqn:kitchen}
\CornerResponse_K&=\frac{g_{xx}g_y^2 + g_{yy}g_x^2 - 2g_{xy}g_xg_y}{g_x^2g_y^2}\\
\intertext{where, in general,}
g_x &= \pdif{g}{x},\quad g_{xx} = \pdif[2]{g}{x^2}, \quad \text{etc\dots}\notag,
\end{align}
and $g$ is either the image or a bivariate polynomial
fitted locally to the image~\cite{kitchen-rosenfeld-corner}. $\CornerResponse_K$
can also be multiplied by the change in edge direction along the
edge~\cite{singh90corner}. 

Corner strength can also be computed as rate of
change in gradient angle when a bicubic polynomial is fitted to the local image
surface ~\citein{zuniga83facet}{deriche93corner}:
\begin{equation}
\CornerResponse_Z = -2\frac{ c_x^2 c_{y^2} - c_x c_y c_{xy} + c_y^2 c_{x^2}}{(c_x^2 + c_y^2)^\frac{3}{2}},
\end{equation}
where, for example,  $c_{xy}$ is the coefficient of $xy$ in the fitted
polynomial. If edgels are only detected at the steepest part of an edge, then a
score computing total image curvature at the edgels is given by:
\begin{equation}
	\CornerResponse_W = \grad^2\Image - S\abs{\grad\Image}^2.
	\label{eqn:wang}
\end{equation}
where $\grad\Image$ is the image gradient~\cite{wang-brady-corner}.

%
%
\subsubsection{Greylevel derivative based detectors}

The assumption that corners exist along edges is an inadequate model for patches
of texture and point like features, and is difficult to use at junctions.
Therefore a large number of detectors operate directly on greylevel images
without requiring edge detection.

One of the earliest detectors~\cite{beaudet78corner} defines corners to be local
extrema in the determinant of the Hessian:
\begin{equation}
	C_\text{DET} = |\hessian{I}| = I_{xx}I_{yy} - (I_{xy})^2.
\end{equation}
This is frequently referred to as the DET operator.  $C_\text{DET}$ moves along
a line as the scale changes. To counteract this, DET extrema can be found
two scales and connected by a line. Corners are then taken as maxima of the
Laplacian along the line~\cite{deriche91corner}.

Instead of DET maxima, corners can also be taken as the gradient maxima on a
line connecting two nearby points of high Gaussian curvature of opposite sign
where the gradient direction matches the sign change~\cite{dreschler82corner}.
By considering gradients as elementary currents, the magnitude of the
corresponding magnetic vector potential can be computed.  The gradient of this
is taken normal and orthogonal to the local contour direction, and the corner
strength is the multiple of the magnitude of these~\cite{luo98corner}.

\paragraph{Local SSD (Sum of Squared Differences) detectors}

Features can be defined as points with low self-similarity in all directions.
The self-similarity of an image patch can be measured by taking the SSD between
an image patch and a shifted version of itself~\cite{moravec_1980}.  This is the
basis for a large class of detectors.  Harris and Stephens~\cite{Harris88Corner}
built on this by computing an approximation to the second derivative of the SSD
with respect to the shift.  This is both computationally more efficient and can
be made isotropic. The result is:
\def\Harris{\Mat{H}}
\begin{equation}
{
\label{eqn:harrismatrix}
\def\id#1{\pdif{I}{#1}}
\Harris = \left[
\begin{array}{cc}
\widehat{I_x^2} & \widehat{I_xI_y}\\
\widehat{I_xI_y} & \widehat{I_y^2} 
\end{array}
\right],
}
\end{equation}
where $\ \widehat{}\ $ denotes averaging performed over the area of the image
patch.  Because of the wording used in \cite{Harris88Corner}, it is often
mistakenly claimed that \Mat{H} is equal to the negative second derivative of
the autocorrelation. This is not the case because the SSD is equal to the sum of
the autocorrelation and some additional terms~\cite{rosten_2006_thesis}.

The earlier F\"{o}rstner~\cite{forstner86feature} algorithm is easily easily explained in
terms of \Harris. For a more recently proposed detector~\cite{tomasi91corner},
it has been shown 
shown~\cite{shi93good} that under affine motion, it is better to use the
smallest eigenvalue of \Harris as the corner strength function.
A number of other suggestions
\cite{Harris88Corner,noble-descriptions,shi93good,Kenney03-condition} have been
made for how to compute the corner strength from \Harris, and these have been
shown to all be equivalent to various matrix norms of
\Harris~\cite{Zuliani04-MathematicalComparison}.  \Harris can be generalized by
generalizing the number of channels and dimensionality of the
image~\cite{kenney05axiomatic} and it can also be shown that that
\cite{forstner86feature,shi93good,rohr97differential} are equivalent to specific
choices of the measure used in \cite{Kenney03-condition}.

\Harris can be explained in terms of the first fundamental form of the image
surface~\cite{noble-corners}. From analysis of the second fundamental form, a
new detector is proposed which detects points where the probability of the
surface being hyperbolic is high.

Instead of local SSD, general template matching, given a warp, appearance model
and pointwise comparison which behaves similarly to the SSD (sum of squared
differences) for small differences can be considered~\cite{triggs04keypoints}.
The stability with respect to the match parameters is derived, and the result is
a generalization of \Harris (where \Harris is maximally stable for no appearance
model, linear translation and SSD matching). This is used to derive detectors
which will give points maximally stable for template matching, given similarity
transforms, illumination models and prefiltering.

%
%
\paragraph{Laplacian based detectors}

An alternative approach to the problem of finding a scalar value which measures
the amount of second derivative is to take the Laplacian of the image.  Since
second derivatives greatly amplify noise, the noise is reduced by using the
smoothed Laplacian, which is computed by convolving the image with the LoG
(Laplacian of a Gaussian).  Since the LoG kernel is symmetric, one can interpret
this as performing matched filtering for features which are the same shape as a
LoG. As a result, the variance of the Gaussian determines the size of features
of interest.  It has been noted~\cite{affine-schmid} that the locations of
maxima of the LoG over different scales are particularly stable.

Scale invariant corners can be extracted by convolving the image with a DoG
(Difference of Gaussians) kernel at a variety of scales (three per octave) and
selecting local maxima in space and scale~\cite{SIFT}.  DoG is a good
approximation for LoG and is much faster to compute, especially as the
intermediate results are useful for further processing.  To reject edge-like
features, the eigenvalues of the Hessian of the image are computed and features
are kept if the eigenvalues are sufficiently similar (within a factor of 10).
This method can be contrasted with \Eqn{wang}, where the Laplacian is compared
to the magnitude of the edge response.  If two scales per octave are
satisfactory, then a significant speed increase can be achieved by using
recursive filters to approximate Gaussian convolution~\cite{crowley03fast}.

Harris-Laplace~\cite{harris-laplace} features are detected using a similar
approach. An image pyramid is built 
and features are detected by computing
$\CornerResponse_H$  at each layer of the pyramid.  Features are
selected if they are a local maximum of  $\CornerResponse_H$ in the image plane
and a local maxima of the LoG across scales.

Recently, scale invariance has been extended to consider features which are
invariant to affine transformations
\cite{affine-schmid,lowe-brown-affine,affine_features,zisserman-affine}.
However, unlike the 3D scale space, the 6D affine space is too large to search,
so all of these detectors start from corners detected in scale space. These in
turn rely on 2D features selected in the layers of an image pyramid.

%
%
\subsubsection{Direct greylevel detectors}

Another major class of corner detectors work by examining a small patch of an
image to see if it ``looks'' like a corner. The detectors described in this
paper belong in this section.

\paragraph{Wedge model detectors}

A number of techniques assume that a corner has the general appearance of one or
more wedges of a uniform intensity on a background of a different uniform
intensity.   For instance a corner can be modelled as a
single~\cite{guiducci88corner} or family~\cite{rohr92corner} of blurred wedges
where the parameters are found by fitting a parametric model. The model can
include angle, orientation, contrast, bluntness and curvature of a single
wedge~\cite{rosin99corner}.  In a manner similar to \cite{canny}, convolution
masks can be derived for various wedges which optimize signal to noise ratio and
localisation error, under assumption that the image is corrupted by Gaussian
noise~\cite{rangarajan88corner}.

It is more straightforward to detect wedges in binary images and to get useful
results, local thresholding can be used to binarize the
image~\cite{s-liu90corner}. If a corner is a bilevel wedge, then a response
function based on local Zernike moments can be used to detect
corners~\cite{ghosal94corner}. A more direct method for finding wedges is to
find points where  where concentric contiguous arcs of pixels are significantly
different from the centre pixel~\cite{shen00real}. According to the wedge model,
a corner will be the intersection of several edges.  An angle-only
Hough transform~\cite{duda72hough} is performed on edgels belonging to lines passing
through a candidate point to find their angles and hence detect
corners~\cite{shen02corner}. Similar reasoning can be used to derive a response
function based on gradient moments to detect V-, T- and X- shaped
corners~\cite{luo04corner}.  The strength of the edgels, wedge angle and
dissimilarity of the wedge regions has also been used to find
corners~\cite{xie93corner}.

\paragraph{Self dissimilarity}

The tip of a wedge is not self-similar, so this can be generalized by defining
corners as points which are not self-similar.  The proportion of pixels in a
disc around a centre (or \term{nucleus}) which are similar to the centre is a
measure of self similarity. This is the USAN (univalue segment assimilating
nucleus).  Corners are defined as SUSAN (smallest USAN, \ie local minima) points
which also pass a set of rules to suppress qualitatively bad features. In
practice, a weighted sum of the number of pixels inside a disc whose intensity
is within some threshold of the centre value is used~\cite{Smith97SUSAN}.
COP (Crosses as Oriented Pair) \cite{bae2002cop} computes dominant
directions using local averages USANs of a pair of oriented crosses, and define
corners as points with multiple dominant directions.

Self similarity can be measured using  a circle instead of a
disc~\cite{Trajkovic-Headley-fast}. The SSD between the center pixel and the
pixels at either end of a diameter line is an oriented measure of
self-dissimilarity. If this is small in any orientation then the point is not a
corner. This is computationally efficient since the process can be stopped as
soon as one small value is encountered. This detector is also used
by~\cite{LepetitF06} with the additional step that the difference between the
centre pixel and circle pixels is used to estimate the Laplacian, and points are
also required to be locally maximal in the Laplacian.

Small regions with a large range in greyvalues can be used as corners. To find
these efficiently, the image can be projected on to the $x$ and $y$ axes and
large peaks found in the second derivatives. Candidate corner locations are the
intersections of these maxima projected back in to the image~\cite{wu83corner}.
Paler \etal~\cite{paler84corner} proposes self similarity can be measured by
comparing the centre pixel of a window to the median value of pixels in the
window.  In practice, several percentile values (as opposed to just the
50$^\text{th}$) are used.

Self-dissimilar patches will have a high energy content. Composing two
orthogonal quadrature pair Gabor filters gives oriented energy.  Corners are
maxima of total energy (the sum of oriented energy over a number of
directions)~\cite{robbins97feature}. 

A fast radial symmetry transform is developed in \cite{loy02fast} to detect
points. Points have a high score when the gradient is both radially symmetric,
strong, and of a uniform sign along the radius.  The detected points have some
resemblance DoG features.

\paragraph{Machine learning based detectors}

All the detectors described above define corners using a model or algorithm and
apply that algorithm directly to the image.  An alternative is to train a
classifier on the model and then apply the classifier to the image.  For
instance, a multilayer perception can be trained on example corners from some
model and applied to the image after some processing
\cite{dias95neural1,chen97corner}.

Human perception can be used instead of a model~\cite{kienzle05interest}:
images are shown to a number of test subjects. Image locations
which are consistently fixated on (as measured by an eye tracking system) are
taken to be interesting, and a support vector machine is trained to recognize
these points.

If a classifier is used, then it can be trained according to how a corner should
behave, \ie that its performance in a system for evaluating detectors should be
maximized.  Trujillo and Olague \cite{trujillo06synthesis} state that detected
points should have a high repeatability (as defined by
\cite{schmid2000comparing}), be scattered uniformly across the image and that
there should be at least as many points detected as requested.  A corner
detector function is optimized (using genetic programming) to maximize the score
based on these measures.

The FAST-$n$ detector (described in \Sec{st}) is related to the
wedge-model style of detector evaluated using a circle surrounding the candidate
pixel.  To optimize the detector for speed, this model is used to train a
decision tree classifier and the classifier is applied to the image.  The
FAST-ER detector (described in \Sec{FASTER}) is a generalization which allows
the detector to be optimized for repeatability.

\subsection{Comparison of feature detectors}

Considerably less work has been done on comparison and evaluation of feature
detectors than on inventing new detectors.  The tests fall into three broad
categories\footnote{Tests for the localisation accuracy are not considered here since
for most applications the presence or absence of useful corners is the limiting
factor}: \begin{enumerate}

\item {\em Corner detection as object recognition.} Since there is no good
definition of exactly what a corner should look like, algorithms can be compared
using simplistic test images where the performance is evaluated (in terms of
true positives, false positives, etc\ldots) as the image is altered using
contrast reduction, warps and added noise.  Since a synthetic image is used,
corners exist only at known locations, so the existence of false negatives and
false positives is well defined. However, the method and results do not
generalize to natural images. 

\item {\em System performance.}
The performance of an application (often tracking) is evaluated as the corner
detector is changed. The advantage is that it tests the suitability of detected
corners for further processing.  However, poor results would be obtained from a
detector ill matched to the downstream processing. Furthermore the results do
not necessarily generalize well to other systems. To counter this, sometimes
part of a system is used, though in this case the results do not necessarily
apply to any system.

\item {\em Repeatability.} This tests whether corners are detected from multiple
views. It is a low level measure of corner detector quality and provides an
upper bound on performance. Since it is independent of downstream processing,
the results are widely applicable, but it is possible that the detected features
may not be useful. Care must be used in this technique, since the trivial
detector which identifies every pixel as a corner achieves 100\% repeatability.
Furthermore, the repeatability does not provide information about the
usefulness of the detected corners for further processing. For instance,
the brightest pixels in the image are likely to be repeatable but not especially
useful.
\end{enumerate}

In the first category, Rajan and Davidson~\cite{rajan89corner} produce a number
of elementary test images with a very small number of corners (1 to 4) to test
the performance of detectors as various parameters are varied. The parameters
are corner angle, corner arm length, corner adjacency, corner sharpness,
contrast and additive noise. The positions of detected corners are tabulated
against the actual corner positions as the parameters are varied.  Cooper
\etal~\cite{cooper91corner,cooper-corner} use a synthetic test image consisting
of regions of uniform intensity arranges to create L-, T-, Y-  and X-shaped
corners. The pattern is repeated several times with decreasing contrast.
Finally, the image is blurred and Gaussian noise is added.  Chen
\etal~\cite{chen97corner} use a related method. A known test pattern is
subjected to a number random affine warps and contrast changes. They note that
this is na\"\i{}ve, but tractable.  They also provide an equivalent to the ROC
(Receiver Operating Characteristic) curve.  Zhang \etal~\cite{zhang94corner2}
generate random corners according to their model and plot localization error,
false positive rate and false negative rate against the detector and generated
corner parameters.  Luo \etal~\cite{luo98corner} use an image of a carefully
constructed scene and plot the proportion of true positives as the scale is
varied and noise is added for various corner angles.

Mohanna and Mokhtarian~\cite{mohanna-performance} evaluate
performance using several criteria. Firstly, they define a {\it consistent}
detector as one where the number of detected corners does not vary with various
transforms such as addition of noise and affine warping. This is measured by the
`consistency of corner numbers' (CCN):
\begin{equation}
CCN = 100 \times 1.1^{-|n_t - n_o|},
\end{equation}
where $n_t$ is the number of features in the transformed image and $n_o$ is the
number of features in the original image.  This test does not determine the
quality of the detected corners in any way, so they also propose measuring the
accuracy (ACU) as:
\begin{equation}
ACU = 100 \times \frac{\frac{n_a}{n_o} + \frac{n_a}{n_g}}{2},
\end{equation}
where $n_o$ is the number of detected corners, $n_g$ is the number of so-called
`ground truth' corners and $n_a$ is the number of detected corners which are
close to ground truth corners. Since real images are used, there is no good
definition of ground truth, so a number of human test subjects (\eg 10)
familiar with corner detection in general, but not the methods under test,
label corners in the test images. Corners which 70\% of the test subjects agree
on are kept as ground truth corners.  This method unfortunately relies on
subjective decisions.

Remarkably, of the systems above, only
\cite{chen97corner}, \cite{rajan89corner} and \cite{kienzle05interest}
provide ROC curves (or equivalent): otherwise only a single point (without
consistency on either axis of the ROC graph) is measured.

In the second category, Trajkovic and Hedley~\cite{Trajkovic-Headley-fast}
define stability as the number of `strong' matches, matches detected over
three frames in their tracking algorithm, divided by the total number of
corners.  Tissainayagama and Suterb~\cite{assess-peformance-corner} use a
similar method: a corner in frame $n$ is stable if it has been successfully
tracked from frame 1 to frame $n$. Bae \etal~\cite{bae2002cop} detect optical
flow using cross correlation to match corners between frames and compare the
number of matched corners in each frame to the number of corners in the first
frame.  

To get more general results than provided by system performance, the performance
can be computed using only one part of a system.  For instance, Mikolajczyk and
Schmid \cite{schmid-2005-performance} test a large number of interest point
descriptors and a small number of closely related detectors by computing how
accurately interest point matching can be performed. Moreels and
Perona~\cite{moreels07evaluation} perform detection and matching experiments
across a variety of scene types under a variety of lighting conditions. Their
results illustrate the difficulties in generalizing from system performance
since the best detector varies with both the choice of descriptor and lighting
conditions.

In the third category, Schmid \etal~\cite{schmid2000comparing} propose that when
measuring reliability, the important factor is whether the same real-world
features are detected from multiple views.  For an image pair, a feature is
`detected' if it is extracted in one image and appears in the second. It is
`repeated' if it is also detected nearby in the second. The repeatability is the
ratio of repeated features to detected features.  They perform the tests on
images of planar scenes so that the relationship between point positions is a
homography.  Fiducial markers are projected onto the planar scene using an
overhead projector to allow accurate computation of the homography.
To measure the suitability of interest points for further processing, the
information content of descriptors of patches surrounding detected points is
also computed.

\section{High-speed corner detection}
\label{sec:st}
\subsection{FAST: Features from Accelerated Segment Test}

\makefig[0.48]{corner/corner.eps}{tomcorner}{ 12 point segment test corner
detection in an image patch.  The highlighted squares are the pixels used in
the corner detection. The pixel at $p$ is  the centre of a candidate corner.
The arc is indicated by the dashed line passes through 12 contiguous pixels
which are brighter than $p$ by more than the threshold.}{}

The segment test criterion operates by considering a circle of sixteen pixels
around the corner candidate $p$.  The original detector
\cite{corner-tracking-paper,label-placement-paper} classifies $p$ as a corner
if there exists a set of $n$ contiguous pixels in the circle which are all
brighter than the intensity of the candidate pixel $I_p$ plus a threshold $t$,
or all darker than $I_p-t$, as illustrated in \Fig{tomcorner}.  $n$ was
originally chosen to be twelve because it admits a high-speed test which can be
used to exclude a very large number of non-corners.  The high-speed test examines 
pixels 1 and 9. If both of these are within $t$ if $I_p$, then $p$ can not be a
corner. If $p$ can still be a corner, pixels 5 and 13 are examined.
If $p$ is a
corner then at least three of these must all be brighter than $I_p+t$ or darker
than $I_p-t$.  If neither of these is the case, then $p$ cannot be a corner.
The full segment test criterion can then be applied to the remaining candidates
by examining all pixels in the circle.  This detector in itself exhibits high
performance, but there are several weaknesses:
\begin{enumerate}
\item This high-speed test does not reject as many candidates for $n < 12$, since the
point can be a corner if only two out of the four pixels are both significantly brighter or
both significantly darker than $p$ (assuming the pixels are adjacent).
Additional tests are also required to find if the complete test needs to be
performed for a bright ring or a dark ring.

\item The efficiency of the detector will depend on the ordering of the
questions and the distribution of corner appearances. It is unlikely that this
choice of pixels is optimal.

\item Multiple features are detected adjacent to one another.
\end{enumerate}

\subsection{Improving generality and speed with machine learning}
\label{sec:learntree}

Here we expand on the work first presented in \cite{fast-paper} and present an
approach which uses machine learning to address the first two points (the
third is addressed in \Sec{nonmax}). 
The process operates in two stages. First, to build a corner detector for a
given $n$, all of the 16 pixel rings are extracted a set of images (preferably
from the target application domain). These are labelled using a
straightforward implementation of the segment test
criterion for $n$ and a convenient threshold. 

\def\px{{p\rightarrow x}}

For each location on the circle $x \in \{1\ldots16\}$, the pixel at that position
relative to $p$, denoted by 
$p \rightarrow x$, can have one of three states: 
\begin{equation}
S_{p\rightarrow x} = \left\{ \begin{array} {lrcll}
d, & &I_{p\rightarrow x}& \le I_p-t				& \text{(darker)}\\
s, & \quad I_p - t <& I_{p\rightarrow x} &< I_p +t\quad& \text{(similar)}\\
b, & I_p + t \le& I_{p\rightarrow x}&			&	\text{(brighter)}		
\end{array} \right.\label{eqn:partition}
\end{equation} Let $P$ be the set of all pixels in all training images.
Choosing an $x$ partitions $P$ into three subsets,  $P_d$, $P_s$ and $P_b$,
where:
\begin{equation}
P_b = \{ p \in P : S_{p \rightarrow x} = b \},
\end{equation}
and $P_d$ and $P_s$ are defined similarly.
In other words, a given choice of $x$ is used to partition the data in to three
sets. The set $P_d$ contains all points where pixel $x$ is darker than the center pixel
by a threshold $t$, $P_b$ contains points brighter than the centre pixel by $t$,
and $P_s$ contains the remaining points 
where pixel $x$ is similar to the centre pixel.

Let $K_p$ be a boolean variable which is true if $p$ is a corner and false otherwise.
Stage 2 employs the algorithm used in ID3~\cite{id3} and begins by selecting
the $x$ which yields the most information about whether the candidate
pixel is a corner, measured by the entropy of $K_p$.

\def\cbar{\ensuremath{\bar{c}}\xspace}

The total entropy of $K$ for an arbitrary set of corners, $Q$, is:
\begin{align}
H(Q) &= (c+\cbar)\log_2(c+\cbar) - c\log_2c - \cbar\log_2 \cbar\\
\text{where}\qquad
c    &= \big|\{i \in Q: K_i \text{ is true}\}\big|\qquad\text{(number of corners)}&\notag\\
\text{and}\qquad
\cbar &= \big|\{i \in Q: K_i \text{ is false}\}\big|\qquad\text{(number of non corners)}\notag&
\end{align}
The choice of $x$ then yields the information gain ($H_g$):
\begin{equation}
H_g = H(P) - H(P_d) - H(P_s) - H(P_b)
\end{equation}
Having selected the $x$ which yields the most information, the process is
applied recursively on all three subsets
\ie $x_b$ is selected to partition $P_b$ in to $P_{b,d}$, $P_{b,s}$, $P_{b,b}$, 
$x_s$ is selected to partition $P_s$ in to $P_{s,d}$, $P_{s,s}$, $P_{s,b}$ and
so on, where each $x$ is chosen to yield maximum information about the set it is
applied to. 
The recursion process terminates when the entropy of a subset is zero.  This
means that all $p$ in this subset have the same value of $K_p$, \ie they are
either all corners or all non-corners.  This is guaranteed to occur since $K$ is
an exact function of the data. In summary, this procedure creates a decision
tree which can correctly classify all corners seen in the training set and
therefore (to a close approximation) correctly embodies the rules of the chosen
FAST corner detector.

In some cases, two of the three subtrees may be the same. In this case, the
boolean test which separates them is removed.  This decision tree is then
converted into C code, creating a long string of nested if-else statements
which is compiled and used as a corner detector.  For highest speed operation,
the code is compiled using profile guided optimizations which allow branch
prediction and block reordering optimizations. 

For further optimization, we force $x_b$, $x_d$ and $x_s$ to be equal. In this
case, the second pixel tested is always the same. Since this is the case, the
test against the first and second pixels can be performed in batch.  This allows
the first two tests to be performed in parallel for a strip of pixels using the
vectorizing instructions present on many high performance microprocessors. Since
most points are rejected after two tests, this leads to a significant speed
increase.

Note that since the data contains incomplete coverage of all possible corners,
the learned detector is not precisely the same as the segment test detector.
In the case of the FAST-$n$ detectors, it is straightforward to include an
instance of every possible combination of pixels (there are $3^{16} =
43,046,721$ combinations) with a low weight to ensure that the learned detector
exactly computes the segment test cirterion.

\subsection{Non-maximal suppression}
\label{sec:nonmax}

Since the segment test does not compute a corner response function, non maximal
suppression can not be applied directly to the resulting features.
For a given $n$, as $t$ is increased, the number of detected corners will
decrease.  Since $n=9$ produces the best repeatability results (see
\Sec{results}), variations in $n$ will not be considered.
The corner strength is therefore defined to be the maximum value of
$t$ for which a point is detected as a corner.


The decision tree classifier can efficiently determine the class of a pixel for
a given value of $t$.  The class of a pixel (for example, 1 for a corner, 0 for
a non-corner) is a monotonically decreasing function of $t$. Therefore, we can
use bisection to efficiently find the point where the function changes from 1
to 0.  This point gives us the largest value of $t$ for which the point is
detected as a corner. Since $t$ is discrete, this is the binary search
algorithm.

Alternatively, an iteration scheme can be used. A pixel on the ring `passes' the
segment test if it is not within $t$ of the centre. If enough pixels fail, then
the point will not be classified as a corner. The detector is run, and of all
the pixels which pass the test, the \emph{amount} by which they pass is found.
The threshold is then increased by the smallest of these amounts, and the
detector is rerun. This increases the threshold just enough to ensure
that a different path is taken through the tree. This process is then iterated
until detection fails.

Because the speed depends strongly on the learned tree and the specific
processor architecture, neither technique has a definitive speed advantage over
the other.  Non maximal suppression is performed in a $3\times3$ mask.

\section{Measuring detector repeatability}
\label{sec:repeatability}

For an image pair, a feature is `useful' if it is extracted in one image and can
potentially appear in the second (i.e.\ it is not occluded). It is `repeated' if
it is also detected nearby the same real world point in the second. For the
purposes of measuring repeatability this allows several features in the first
image to match a single feature in the second image.
The repeatability, $R$, is defined to be
\begin{equation}
\label{eqn:repeatability}
R = \frac{N_\text{repeated}}{N_\text{useful}},
\end{equation}
where $N_\text{repeated}$ and  $N_\text{useful}$ are summed over all image pairs
in an image sequence. This is equivalent to the weighted average of the
repeatabilities for each image pair, where the weighting is the number of
useful features. In this paper, we generally compute the repeatability for a
given number of features per frame, varying between zero and 2000 features (for
a $640\times480$ image). This also allows us to compute the area under the
repeatability curve, $A$, as an aggregate score.

\makefig[0.45]{repeatability-explain/repeatability}{repeat-example}{Repeatability
is tested by checking if the same real-world features are detected in different
views. A geometric model is used to compute where the features reproject to.}{}

The repeatability measurement requires the location and visibility of every
pixel in the first image to be known in the second image.  In order to compute
this, we use a 3D surface model of the scene to compute if and where where
detected features should appear in other views. This is illustrated in
\Fig{repeat-example}.  This allows the repeatability of the detectors to be
analysed on features caused by geometry such as corners of polyhedra, occlusions
and junctions.  We also allow bas-relief textures to be modelled with a flat
plane so that the repeatability can be tested under non-affine warping.

The definition of `nearby' above must allow a small margin of error
($\varepsilon$ pixels) because the alignment, the 3D model and the camera
calibration (especially the radial distortion) is not perfect. Furthermore, the
detector may find a maximum on a slightly different part of the corner. This
becomes more likely as the change in viewpoint and hence change in shape of the
corner become large.  

Instead of using fiducial markers, the 3D model is aligned to the scene by hand
and this is then optimised using a blend of simulated annealing and gradient
descent to minimise the SSD (sum of squared differences) between all pairs of
frames and reprojections.
To compute the SSD between frame $i$ and reprojected frame $j$, the position of
all points in frame $j$ are found in frame $i$. The images are then bandpass
filtered. High frequencies are removed to reduce noise, while low frequencies
are removed to reduce the impact of lighting changes.  To improve the speed of
the system, the SSD is only computed using 1000 random locations.

The datasets used are shown in \Fig{box}, \Fig{maze4} and \Fig{junk}. With
these datasets, we have tried to capture a wide range of geometric and textural
corner types.

\begin{figure}[t]
\def\file{repeatability-datasets/box/video_frame-00}
\def\figw{0.10\textwidth}
\centering
\Includegraphics[width=\figw]{\file 00.eps}
\Includegraphics[width=\figw]{\file 01.eps}
\Includegraphics[width=\figw]{\file 02.eps}
\Includegraphics[width=\figw]{\file 03.eps}
\Includegraphics[width=\figw]{\file 04.eps}
\Includegraphics[width=\figw]{\file 05.eps}
\Includegraphics[width=\figw]{\file 06.eps}
\Includegraphics[width=\figw]{\file 07.eps}
\Includegraphics[width=\figw]{\file 08.eps}
\Includegraphics[width=\figw]{\file 09.eps}
\Includegraphics[width=\figw]{\file 10.eps}
\Includegraphics[width=\figw]{\file 11.eps}
\Includegraphics[width=\figw]{\file 12.eps}
\Includegraphics[width=\figw]{\file 13.eps}
\caption{\label{fig:box}Box dataset: photographs taken of a test rig
(consisting of photographs pasted to the inside of a cuboid) with strong
changes of perspective, changes in scale and large amounts of radial
distortion. This tests the corner detectors on planar texture.} 
\end{figure}
\begin{figure}[t]
\def\file{repeatability-datasets/maze-4mm/video_frame-00}
\def\figw{0.10\textwidth}
\centering
\Includegraphics[width=\figw]{\file 00.eps}
\Includegraphics[width=\figw]{\file 01.eps}
\Includegraphics[width=\figw]{\file 02.eps}
\Includegraphics[width=\figw]{\file 03.eps}
\Includegraphics[width=\figw]{\file 04.eps}
\Includegraphics[width=\figw]{\file 05.eps}
\Includegraphics[width=\figw]{\file 06.eps}
\Includegraphics[width=\figw]{\file 07.eps}
\Includegraphics[width=\figw]{\file 08.eps}
\Includegraphics[width=\figw]{\file 09.eps}
\Includegraphics[width=\figw]{\file 10.eps}
\Includegraphics[width=\figw]{\file 11.eps}
\Includegraphics[width=\figw]{\file 12.eps}
\Includegraphics[width=\figw]{\file 13.eps}
\Includegraphics[width=\figw]{\file 14.eps}
\caption{\label{fig:maze4}Maze dataset: photographs taken of a prop used in an
augmented reality application. This set consists of textural features undergoing
projective warps as well as geometric features. There are also significant
changes of scale.}
\end{figure}
\begin{figure}[t]
\def\file{repeatability-datasets/junk/video_frame-00}
\def\figw{0.10\textwidth}
\centering
\Includegraphics[width=\figw]{\file 00.eps}
\Includegraphics[width=\figw]{\file 01.eps}
\Includegraphics[width=\figw]{\file 02.eps}
\Includegraphics[width=\figw]{\file 03.eps}
\Includegraphics[width=\figw]{\file 04.eps}
\Includegraphics[width=\figw]{\file 05.eps}
\Includegraphics[width=\figw]{\file 06.eps}
\Includegraphics[width=\figw]{\file 07.eps}\\
\caption{\label{fig:junk}Bas-relief dataset: the model is a flat plane, but
there are many objects with significant relief. This causes the appearance of
features to change in a non affine way from different viewpoints.
}
\end{figure}

\section{FAST-ER: Enhanced Repeatability}
\label{sec:FASTER}

Since the segment test detector can be represented as a ternary decision tree
and we have defined repeatability, the detector can be generalized by defining a
feature detector to be a ternary decision tree which detects points with high
repeatability. The repeatability of such a detector is a non-convex function of
the configuration of the tree, so we optimize the tree using simulated
annealing.  This results in a multi-objective optimization. If every point is
detected as a feature, then the repeatability is trivially perfect. Also, if the
tree complexity is allowed to grow without bound, then the optimization is quite
capable of finding one single feature in each image in the training set which
happens to be repeated. Neither of these are useful results.  To account for
this, the cost function for the tree is defined to be:
\begin{equation}
	k = 
		\left(1 + \left(\frac{w_r}{r}\right)^2 \right) 
		\left(1 + \frac{1}{N}\sum_{i=1}^N\left(\frac{{d_i}}{w_n}\right)^2 \right) 
		\left(1 + \left(\frac{s}{w_s}\right)^2 \right),
\end{equation}
where $r$ is the repeatability (as defined in \Eqn{repeatability}), ${d_i}$ is
the number of detected corners in frame $i$, $N$ is the number of frames and
$s$ is the size (number of nodes) of the decision tree.  The effect of these
costs are controlled by $w_r$, $w_n$, and $w_s$. Note that for efficiency,
repeatability is computed at a fixed threshold as opposed to a fixed number of
features per frame.

The corner detector should be invariant to rotation, reflection and intensity
inversion of the image.
To prevent excessive burden on the optimization algorithm, each time the tree is
evaluated, it is applied sixteen times: at four rotations, 90\degrees apart,
with all combinations of reflection and
intensity inversion.  The result
is the logical OR of the detector applications: a corner is detected if any one
of the sixteen applications of the tree classifies the point as a corner.

\begin{figure}
\centering
\Includegraphics[width=0.25\textwidth]{offsets/corner2.eps}
\caption{Positions of offsets used in the FAST-ER detector.\label{fig:newoffsets}}
\end{figure}

Each node of the tree has an offset relative to the centre pixel, $x$, with $x
\in \{0\ldots47\}$ as defined in \Fig{newoffsets}. Therefore, $x=0$ refers to the
offset $(-1,4)$.  Each leaf has a class $K$, with 0 for non-corners and 1 for
corners.  Apart from the root node, each node is either on a $b$, $d$ or $s$
branch of its parent, depending on the test outcome which leads to that branch. The
tree is constrained so that each leaf on an $s$ branch of its direct parent has
$K=0$.  This ensures that the number of corners generally decreases as the
threshold is increased. 

The simulated annealing optimizer makes random
modifications to the tree by first selecting a node at random and then mutating
it. If the selected node is:
\begin{itemize}
\item a leaf, then with equal probability, either:
	\begin{enumerate}
		\item Replace node with a random subtree of depth 1. 
		\item Flip classification of node. This choice is not available if the
		      leaf class is constrained.
	\end{enumerate}
\item a node, then with equal probability, choose any one of:
	\begin{enumerate}
		\item Replace the offset with a random value in ${0\ldots47}$.
		\item Replace the node with a leaf with a random class (subject to the
		constraint).
		\item Remove a randomly selected branch of the node and replace it with
		a copy of another randomly selected branch of that node. For example, a $b$ branch
		may be replaced with a copy of an $s$ branch.
	\end{enumerate}
\end{itemize}
The randomly grown subtree  consists of a single decision node (with a random
offset in ${0\ldots47}$), and three leaf nodes.  With the exception of the
constrained leaf, the leaves of this random subtree have random classes. 
These modifications to
the tree allow growing,
mutation, mutation and shrinking of the tree, respectively. The last modification of the tree
is motivated by our observations of the FAST-9 detector. 
In FAST-9, a large number of nodes have the characteristic that two out of the
three subtrees are identical.
Since FAST-9 exhibits high repeatability, we have included this modification to
allow FAST-ER to easily learn a similar structure.

The modifications are accepted according to the Boltzmann acceptance criterion,
where the probability $P$ of accepting a change at iteration $I$ is:
\begin{equation}
P = e^{\frac{\hat{k}_{I-1} - k_I}{T}}
\end{equation}
where $\hat{k}$ is the cost after application of the acceptance criterion and
$T$ is the temperature. The temperature follows an exponential schedule:
\begin{equation}
T = \beta e^{-\alpha\frac{I}{I_\text{max}}},
\end{equation}
where $I_\text{max}$ is the number of iterations. The algorithm is initialized
with a randomly grown tree of depth 1, and the algorithm uses a fixed threshold,
$t$. Instead of performing a single optimization, the optimizer is rerun a
number of times using different random seeds.

Because the detector must be applied to
the images every iteration, each candidate tree in all sixteen transformations
is compiled to machine code in memory and executed directly. 
Since it is applied with sixteen transformations, the
resulting detector is not especially efficient. So for efficiency, the
detector is used to generate training data  so that a single tree can be
generated using the method described in \Sec{learntree}. The resulting tree
contains approximately 30,000 non-leaf nodes.

\subsection{Parameters and justification}

\begin{table}\centering
\begin{tabular}{c|r}
Parameter		&	Value \\\hline
$w_r$			&	    1 	\\
$w_n$			&   3,500	\\
$w_s$			&  	10,000	\\
$\alpha$		&	30		\\
$\beta$			&	100		\\
$t$				&	35		\\
$I_\text{max}$	&	100,000 \\
Runs			&   100		\\
$\varepsilon$   & 5 pixels\\
Training set	&	`box' set, images 0--2.
\end{tabular}
\caption{\label{tab:parameters} Parameters used to optimize the tree.}
\end{table}

The parameters used for training are given in \Tab{parameters}.  The entire
optimization which consists of 100 repeats of a 100,000 iteration optimization
requires about 200 hours on a Pentium 4 at 3GHz.  Finding the optimal set of
parameters is essentially a high dimensional optimization problem, with many
local optima. Furthermore, each evaluation of the cost function is very
expensive. Therefore, the values are in no sense optimal, but they are a set of
values which produce good results.  Refer to \cite{nr_in_C_opt} for techniques
for choosing parameters of a simulated annealing based optimizer.  Recall that the
training set consists of only the first three images from the `box' dataset.

\begin{figure}\centering
\Includegraphics[width=0.45\textwidth]{faster_learn/graph.eps}\\
\caption{\label{fig:FASTERgraph}
Distribution of scores for various parameters of $(w_r, w_n, w_s)$. The
parameters leading to the best result are $(2.0, 3500, 5000)$ and
the parameters for the worst point are $(0.5, 3500, 5000)$. For comparison, the
distribution for all 27 runs and the median point (given in \Tab{parameters})
are given.  The score given is the mean value of $A$ computed over the `box',
`maze' and `bas-relief' datasets.
}
\end{figure}

The weights determine the relative effects of good repeatability, resistance to
overfitting and corner density, and therefore will affect the performance of
the resulting corner detector.  To demonstrate the sensitivity of the detector
with respect to $w_r$, $w_n$ and $w_s$ a detector was learned for three
different values of each, $w_r \in \{0.5, 1, 2\}$, $w_n \in \{1750, 5300, 7000\}$
and $w_s \in \{5000, 10000, 20000\}$, resulting in a total of 27 parameter
combinations.
The performance of the detectors
are evaluated by computing the mean area under the repeatability curve for the
`box', `maze' and `bas-relief' datasets. Since in each of the 27 points, 100
runs of the optimization are performed, each of the 27 points produces a
distribution of scores. The results of this are shown in \Fig{FASTERgraph}. The
variation in score with respect to the parameters is quite low even though the
parameters all vary by a factor of four. Given that, the results for the set of
parameters in \Tab{parameters}  are very close to the results for the best
tested set of parameters. This demonstrates that the choices given in
\Tab{parameters} are reasonable.

\section{Results}
\label{sec:results}

In this section, the FAST and FAST-ER detectors are compared against a variety
of other detectors both in terms of repeatability and speed. In order to test
the detectors further, we have used the `Oxford' dataset
~\cite{oxford-vgg-dataset} in addition to our own.
This dataset models the warp between images using a homography, and consists of
eight sequences of six images each. It tests detector repeatability under
viewpoint changes (for approximately planar scenes), lighting changes, blur and
JPEG compression.
Note that the FAST-ER detector is trained on 3 images
(6 image pairs), and is tested on a total of 85 images (688 image pairs).

The parameters used in the various detectors are given in \Tab{testparams}. In
all cases (except SUSAN, which uses the reference implementation in
\cite{SUSANcode}), non-maximal suppression is performed using a $3\times3$
mask.
 The number of features was controlled in a manner equivalent to
thresholding on the response.
For the Harris-Laplace detector, the Harris
response was used, and for the SUSAN detector, the `distance threshold'
parameter was used.
It should be noted that some experimentation was performed on all the detectors
to find the best results on our dataset. In the case of FAST-ER, the
best detector was selected. The parameters were then used without modification
on the `Oxford' dataset.  The timing results were obtained with the same
parameters used  in the repeatability experiment.

\subsection{Repeatability}

\def\fgraph{\Fig{fastonly}\xspace}
\def\boxgraph{\Fig{camgraphs}{A}\xspace}
\def\mazegraph{\Fig{camgraphs}{B}\xspace}
\def\junkgraph{\Fig{camgraphs}{C}\xspace}
\def\noisegraph{\Fig{noisegraph}\xspace}

\begin{table}\centering
\begin{tabular}[t]{ll}
\bf DoG\\\hline
Scales per octave & 3 \\
Initial blur $\sigma$ & 0.8 \\
Octaves & 4 \\
\\
\bf Harris, Shi-Tomasi\\\hline
Blur $\sigma$ & 2.5 \\
\\
\bf General parameters\\\hline
$\varepsilon$ & 5 pixels
\end{tabular}
\begin{tabular}[t]{ll}
\bf SUSAN\\\hline
Distance threshld & 4.0 \\
\\
\bf Harris-Laplace\\\hline
Initial blur $\sigma$ & 0.8 \\
Harris blur  & 3 \\
Octaves  & 4 \\
Scales per octave & 10\\
\end{tabular}
\caption{\label{tab:testparams}Parameters used for testing corner detectors.}
\end{table}

\begin{figure}
\centering
\Includegraphics[clip,width=0.5\textwidth]{repeatability-results/st/graph.eps}
\caption{\label{fig:fastonly}
A comparison of the FAST-$n$ detectors on the `bas-relief' shows that $n=9$ is the most
repeatable. For $n\le8$, the detector starts to respond strongly to edges.}
\end{figure}

The repeatability is computed as the number of corners per frame is varied. For
comparison we also include a scattering of random points as a baseline measure,
since in the limit if every pixel is detected as a corner, then the
repeatability is 100\%.  To test robustness to image noise, increasing amounts
of Gaussian noise were added to the bas-relief dataset, 
in addition to the significant amounts of camera noise
already present.
Aggregate results taken over all datasets are given in \Tab{AUR}. It can be seen
from this that on average, FAST-ER outperforms all the other tested detectors.

\begin{table}\centering
\begin{tabular}{l|l}
Detector         & $A$\\\hline
FAST-ER          & 1313.6\\
FAST-9           & 1304.57\\
DoG              & 1275.59\\
Shi \& Tomasi    & 1219.08\\
Harris           & 1195.2\\
Harris-Laplace   & 1153.13\\
FAST-12          & 1121.53\\
SUSAN            & 1116.79\\
Random           & 271.73\\
\end{tabular}
\caption{Area under repeatability curves for 0--2000 corners per frame averaged
over all the evaluation datasets (except the additive noise).\label{tab:AUR}}
\end{table}

\begin{figure}\centering
\Includegraphics[clip,width=0.45\textwidth]{results/noise.eps}
\caption{\label{fig:noisegraph}
Repeatability results for the bas-relief data
set (at 500 features per frame) as the amount of Gaussian noise added to the images
is varied. See \Fig{camgraphs} for the key. } 
\end{figure}

More detailed are shown in Figures \ref{fig:fastonly}, \ref{fig:camgraphs} and
\ref{fig:vgggraphs}. As shown in \fgraph, FAST-9 performs best
(FAST-8 and below are edge detectors), so only FAST-9 and FAST-12 (the original
FAST detector) are given.

\begin{figure*}\centering
\begin{tabular}{cc}
\Includegraphics[clip,width=0.45\textwidth]{results/box.eps} &
\Includegraphics[clip,width=0.45\textwidth]{results/maze.eps} \\
\Includegraphics[clip,width=0.45\textwidth]{results/junk.eps} &
\Includegraphics[clip,width=0.20\textwidth]{results/legend.eps}
\end{tabular}
\caption{ \label{fig:camgraphs} \infig{A}, \infig{B}, \infig{C}: Repeatability
results for the repeatability dataset as the number of features per frame is
varied. \infig{D}: Key for this figure, \Fig{vgggraphs} and \Fig{noisegraph}.
For FAST and SUSAN, the number of features can not be chosen arbitrarily; the
closest approximation to 500 features in each frame is used.  }
\end{figure*}

\begin{figure*}\centering
\Includegraphics[clip,width=0.45\textwidth]{results/bark.eps}
\Includegraphics[clip,width=0.45\textwidth]{results/bikes.eps}
\Includegraphics[clip,width=0.45\textwidth]{results/boat.eps}
\Includegraphics[clip,width=0.45\textwidth]{results/graf.eps}
\Includegraphics[clip,width=0.45\textwidth]{results/leuven.eps}
\Includegraphics[clip,width=0.45\textwidth]{results/trees.eps}
\Includegraphics[clip,width=0.45\textwidth]{results/ubc.eps}
\Includegraphics[clip,width=0.45\textwidth]{results/wall.eps}
\caption{\label{fig:vgggraphs}
\infig{A}--\infig{G}:
Repeatability results for the `Oxford' dataset as the number of features
per frame is varied. See \Fig{camgraphs} for the key.}
\end{figure*}

The FAST-9 feature detector, despite being designed only for speed,  generally
outperforms all but FAST-ER on these images.  FAST-$n$, however,
is not very robust to the presence of noise. This is to be expected. High
speed is achieved by analysing the fewest pixels possible, so the detector's
ability to average out noise is reduced. 

The best repeatability results are achieved by FAST-ER.  FAST-ER easily
outperforms FAST-9 in all but Figures~\ref{fig:camgraphs}A,
\ref{fig:vgggraphs}B, C and E. These results are slightly more mixed, but
FAST-ER still performs very well for higer corner densities. FAST-ER greatly
outperforms FAST-9 on the noise test, (and outperforms all other detectors for
$\sigma < 7$).  This is because the training parameters bias the detector
towards detecting more corners for a given threshold than FAST-9.  Consequently,
for a given number of features per frame, the threshold is higher, so the effect
of noise will be reduced.

As the number of corners per frame is increased, all of the detectors, at
some point, suffer from decreasing repeatability. This effect is least
pronounced with the FAST-ER detector. Therefore, with FAST-ER, the corner
density does not need to be as carefully chosen as with the other detectors.
This fall-off is particularly strong in the Harris and Shi-Tomasi detectors.
Shi and Tomasi, derive their result for better feature detection on the
assumption that the deformation of the features is affine.  Their detector
performs slightly better over all, and especially in the cases where the
deformations are largely affine.  For instance, in the bas-relief dataset
(\junkgraph), this assumption does not hold, and interestingly, the Harris
detector outperforms Shi and Tomasi detector in this case. Both of these
detectors tend to outperform all others on repeatability for very low corner
densities (less than 100 corners per frame).

The Harris-Laplace is detector was originally evaluated using planar scenes
\cite{harris-laplace,schmid98comparing}.  he results show that Harris-Laplace
points outperform both DoG points and Harris points in repeatability. For the
box dataset, our results verify that this is correct for up to about 1000 points
per frame (typical numbers, probably commonly used); the results are somewhat
less convincing in the other datasets, where points undergo non-projective
changes.

In the sample implementation of SIFT \cite{sift-code}, approximately 1000 points
are generated on the images from the test sets. We concur that this a good
choice for the number of features since this appears to be roughly where the
repeatability curve for DoG features starts to flatten off.

Smith and Brady\cite{Smith97SUSAN} claim that the SUSAN corner detector performs
well in the presence of noise since it does not compute image derivatives and
hence does not amplify noise. We support this claim. Although the noise results
show that the performance drops quite rapidly with increasing noise to start
with, it soon levels off and outperforms all but the DoG detector. 
The DoG detector is remarkably robust to the presence of noise.  Convolution is
linear, so the computation of DoG is equivalent to convolution with a DoG
kernel. Since this kernel is symmetric, the convolution is equivalent to matched
filtering for objects with that shape. The robustness is achieved because
matched filtering is optimal in the presence of additive Gaussian
noise\cite{some-comms-textbook}.

\subsection{Speed}

Timing tests were performed on a 3.0GHz Pentium 4-D which is representative of a
modern desktop computer. The timing tests are performed on two datasets: the
terst set and the training set.
The training set consists 101 monochrome fields from a high definition video
source with a resolution of $992\times668$ pixels. This video source is used
to train the high speed FAST detectors and for profile-guided optimizations for
all the detectors. The test set consists of 4968 frames of
monochrome $352\times288$ (quarter-PAL) video

The learned FAST-ER, FAST-9 and FAST-12 detectors have been compared to the
original FAST-12 detector, to our implementation of the Harris and DoG
(the detector used by SIFT) and to the reference
implementation of SUSAN\cite{SUSANcode}. The FAST-9, Harris and DoG detectors
use the SSE-2 vectorizing instructions to speed up the processing. The learned
FAST-12 does not, since using SSE-2 does not yield a speed increase.

\begin{table}[tb]
\begin{centering}
\begin{tabular}{l|cc|cc}
Detector                        &  \multicolumn{2}{c|}{Training set} &\multicolumn{2}{c}{Test set}\\
                                &Pixel rate (MPix/s)&  \%           &MPix/s        &   \%      \\\hline
FAST $n=9$                      &    188            &  4.90         &   179        &  5.15     \\
FAST $n=12$ 					&	 158            &  5.88         &   154        &  5.98     \\
Original FAST ($n=12$)          &    79.0           &  11.7         &   82.2       &  11.2     \\
FAST-ER							&    75.4  			&  12.2         &   67.5       &  13.7     \\
SUSAN                           &    12.3           &  74.7         &   13.6       &  67.9     \\
Harris                          &    8.05           &  115          &   7.90       &  117      \\
Shi-Tomasi                      &    6.50           &  142          &   6.50       &  142      \\
DoG                             &    4.72           &  195          &   5.10       &  179      \\
\end{tabular}
\end{centering}\\
\caption{\label{tab:timings}Timing results for a selection of feature detectors
run on frames of two video sequences. The
percentage of the processing budget for $640\times480$ video is given for
comparison.  Note that since PAL, NTSC, DV and 30Hz VGA (common for web-cams)
video have approximately the same pixel rate, the percentages are widely
applicable.  The feature density is equivalent to approximately 500 features per
$640\times480$ frame.  The results shown include the time taken for nonmaximal
suppression.
}
\end{table}

As can be seen in \Tab{timings},  FAST in general is mucxh faster
than the other tested feature detectors, and the learned FAST
is roughly twice as fast as the handwritten version.  In addition, it is
also able to generate an efficient detector for FAST-9, which is the most
reliable of the FAST-$n$ detectors.  Furthermore, it is able to generate a very
efficient detector for FAST-ER. Despite the increased complexity of this
detector, it is still much faster than all but FAST-$n$.
On modern hardware,  FAST and FAST-ER consume only a fraction of the time available during
video processing, and on low power hardware, it is the only one of the detectors
tested which is capable of video rate processing at all.

\section{Conclusions}

In this paper, we have presented the FAST family of detectors. Using machine
learning we turned the simple and very repeatable segment test heuristic into
the FAST-9 detector which has unmatched processing speed. Despite the design for
speed, the resulting detector has excellent repeatability.  By generalizing the
detector and removing preconceived ideas about how a corner should appear, we
were able to optimize a detector directly to improve its repeatability, creating
the FAST-ER detector.  While still being very efficient, FAST-ER has dramatic
improvements in  repeatability over FAST-9 (especially in noisy images). The
result is a detector which is not only computationally efficient, but has better
repeatability results and more consistent with variation in corner density than
any other tested detector.

These results raise an interesting point about corner detection techniques: too
much reliance on intuition can be misleading.  Here, rather than concentrating
on how the algorithm should do its job, we focus our attention on what
performance measure we want to optimize and this yields very good results.  The
result is a detector which compares favourably to existing detectors.

experiment freely available.
The generated FAST-$n$ detectors, the datasets for measuring repeatability, the
FAST-ER learning code and the resulting trees are available from
\footnote{FAST-$n$ detectors are also available in libCVD from:
{\tt
\href{http://savannah.nongnu.org/projects/libcvd}{http://savannah.nongnu.org/projects/libcvd} } }

{\tt
	\href{http://mi.eng.cam.ac.uk/~er258/work/fast.html}{http://mi.eng.cam.ac.uk/~er258/work/fast.html}
}

\clearpage
\bibliographystyle{IEEEtran-ed}
\bibliography{papers}

\end{document}